\documentclass[letterpaper]{article} % DO NOT CHANGE THIS
\usepackage{aaai2026}  % DO NOT CHANGE THIS
\usepackage{times}  % DO NOT CHANGE THIS
\usepackage{helvet}  % DO NOT CHANGE THIS
\usepackage{courier}  % DO NOT CHANGE THIS
\usepackage[hyphens]{url}  % DO NOT CHANGE THIS
\usepackage{graphicx} % DO NOT CHANGE THIS
\usepackage{natbib}  % DO NOT CHANGE THIS AND DO NOT ADD ANY OPTIONS TO IT
\usepackage{caption} % DO NOT CHANGE THIS AND DO NOT ADD ANY OPTIONS TO IT
\usepackage{algorithm}
\usepackage{algorithmic}

\usepackage{newfloat}
\usepackage{listings}
\DeclareCaptionStyle{ruled}{labelfont=normalfont,labelsep=colon,strut=off} % DO NOT CHANGE THIS
\floatstyle{ruled}
\newfloat{listing}{tb}{lst}{}
\floatname{listing}{Listing}
\title{The Benchmark Trap: Structures of Power and Injustice in AI Evaluations}
\author {
 Jason Branford\equalcontrib\textsuperscript{\rm 1},
 Angelie Kraft\equalcontrib\textsuperscript{\rm 2}
}
\affiliations {
 \textsuperscript{\rm 1}University of Hamburg, Hamburg\\
 \textsuperscript{\rm 2}Weizenbaum Institute, Berlin\\
 jason.branford@uni-hamburg.de, angelie.kraft@weizenbaum-institut.de
}

\begin{document}

\maketitle

\begin{abstract}

Artificial intelligence (AI) benchmarks are not neutral tools of evaluation but socio-technical artefacts that shape competition, power, and research priorities within AI. Benchmarks standardise the assessment of systems and facilitate the creation of leaderboards that reward state-of-the-art performance with prestige, citations, trust, and institutional influence. As the costs of developing competitive AI systems rise, these rewards increasingly concentrate among powerful, industry-funded labs. This paper situates these concerns within Iris Marion Young’s theories of oppression and structural injustice. It argues that current benchmarking practices may perpetuate systematic harms affecting various actors in AI research, aligning with four of Young’s ``faces of oppression''. Benchmarking culture is further framed as a source of structural injustice, as these harms emerge from normalised, individually defensible practices and network effects, even without explicit wrongdoing. By reinforcing existing power structures and narrowing possible research trajectories, benchmarking may in fact prevent the field from advancing in epistemically robust and socially beneficial ways.

\end{abstract}

\section{Introduction}
 
Benchmarks do not merely measure the performance of artificial intelligence (AI) systems. They help determine what these systems are taken to be and, therefore, the ends to which they are put, transforming what is inherently a limited technical performance into public evidence of, for example, `reasoning' or `safety'. We assume that many, if not most, individual developers and researchers in the field are genuinely interested in improving benchmarks for the good of the discipline and society. However, AI is not only a field of scientific inquiry but a lucrative business, a (geo)political lever, and a growing object of regulatory concern, which incentivises the involvement of players with different sets of interests. Thus, benchmarks are inherently socio-technical artefacts that are shaped by and contribute to existing power dynamics in AI research. This is especially true of prominent benchmarks that are visible beyond specified academic sub-communities, are broadly utilised to communicate the capabilities of commercial and open-source models towards technical and non-technical audiences, and commonly referenced within AI leaderboards.\footnote{See e.g. \url{https://llm-stats.com/} (access date: July 6, 2026)}

Our central argument is that current benchmarking culture in AI may feed into structures of oppression and injustice towards smaller labs and marginalised researcher communities, on the one side, and towards marginalised user groups, on the other. In so doing, we further suggest that they may therefore be hindering the potential of the field. We draw on the pioneering work of Iris Marion Young~(\citeyear{young-1990-justice, young-2010-responsibility}) to assess different issues in the economy and epistemology of AI benchmarking and conceptualise the kinds of harms that benchmarking perpetuates. After characterising the current culture of benchmarking in AI (Section~\ref{scene}), we detail how stakeholders involved in and affected by the benchmark race may be subject to injustices that align with four of Young's five ``faces of oppression'' (Section~\ref{faces}). We argue that the identified issues with AI benchmarking go beyond the ill-intentioned acts of individual `bad apples' in the AI community but are rather systematic and pose structural harms (Section~\ref{structural}). To this end, we utilise Young's notion of \textit{structural injustice}~(\citeyear{Young_2006, young-2010-responsibility}) and argue, following McKeown~(\citeyear{McKeown_2021, McKeown_2024}), that the types of injustices we identify can be considered \textit{avoidable}. Finally, we consider the implications of these challenges for benchmarking and the future of AI evaluations (Section~\ref{implications}).

\section{The Influence and Troubles of Benchmarks}
\label{scene}

According to the frequently cited definition by~\citet{raji2021ai}, a ``benchmark [is] a particular combination of a dataset or sets of datasets [...], and a metric, conceptualized as representing one or more specific tasks or sets of abilities'' (p. 2). The dataset comes with so-called `ground truth' labels, which distinguish `true' from `false' or `desired' from `undesired' responses. The metric is then used to compute a performance score, which enables the comparison and ranking of systems (e.g., in leaderboards) to determine which fares `best' at a given point in time; noting that a single model is usually evaluated via a variety of benchmarks to assess the extent of its different capabilities. Critical examinations of benchmarks ought to focus on the datasets and metrics used, as well as the ends to which they are put. 

Datasets are a typical focal point, and in the case of AI benchmarks, those used are comparable to AI training datasets in terms of their diversity of content, form, provenance, etc. In order to understand datasets and their societal impacts, it is necessary, as proposed by~\citet{gebru2021datasheets}, to consider both their creators and their users.  \citet{bordes2025evalfactsheetsstructuredframework} identify four scenarios of AI evaluation with different stakeholders: \textit{development-focused} evaluation helps researchers and engineers to analyse model performance and errors quickly and informs design decisions and development approaches; \textit{selection-focused} evaluation guides decision-makers in identifying the right model to deploy for their use case; \textit{deployment-focused} evaluation serves operators and regulators as a measure of ``production readiness'' and the likelihood of ``consequential errors''; \textit{research-focused} evaluation give researchers an insight into what a model is and is not capable of in a more general sense.

These scenarios indicate that benchmarks are not used by a single community for a single purpose, and serve to identify researchers, engineers, decision-makers, operators, and regulators as relevant stakeholders who rely on evaluations in different ways and contexts. Another stakeholder group are journalists reporting on the overall state of AI progress who also draw from (and thereby perpetuate) leaderboards and scores on specific benchmarks.\footnote{\url{https://www.forbes.com/sites/johnkoetsier/2023/03/14/gpt-4-beats-90-of-lawyers-trying-to-pass-the-bar/} (access date: April 26, 2026)} Finally, data authors and annotators are further essential stakeholder groups of AI evaluations, given that the creation of datasets involves significant amounts of human labour, e.g., for writing and annotating test cases~\citep{kraft2025socialbiaspopularquestionanswering}. This work is often carried out under precarious working conditions, characterised by, e.g., low and unstable pay, gig work, isolation, physical and psychological distress~\citep[cf.][]{gebrekidan2024contentmoderation, kapania2026survival}. Moreover, many datasets are based on openly accessible web sources, the scraping of which has been criticised as an extractive practice, often associated with breaches of license agreements and intellectual property rights~\citep[cf.][]{longpre2025bridging}. There are, as such, a variety of stakeholders to be considered when discussing the harms of current benchmarking practices. Benchmarking cannot, therefore, be treated as a narrow methodological concern \textit{internal} to machine learning.

\subsection{Lack of Construct Validity}
\label{validity}
A widely discussed concern around AI benchmarks has been their lack of \textit{construct validity}. ``[C]onstruct validation is involved whenever a test is to be interpreted as a measure of some attribute or quality which is not `operationally defined''', i.e., directly measurable~\citep[][p. 2]{cronbach1955construct}. AI benchmarks are different from rulers or scales in that they indirectly measure abstract constructs for which no definite metric exists. To study this, \citet{wallach2024evaluating} provide an analytical lens that draws from a social scientific framework and distinguishes between a \textit{background concept}, a \textit{systematised concept}, the actual \textit{measurement instrument}, and its resulting \textit{measurements}. A background concept comprises different conceptualisations of a term, for different purposes and by different stakeholders. For instance, software engineers might want to measure a model's mathematical abilities to assess its accuracy in a specific application context. Cognitive scientists, on the other hand, might want to measure mathematical abilities to study similarities and differences to human cognition. \citet{wallach2024evaluating} argue that proper validation of a measurement instrument first requires a systematisation of the concept to be measured. In our example, this entails that a clear-cut definition of `mathematical ability', given a specific purpose and context, must be designated as the systematised concept. The measurement instrument itself can then be designed for and validated against this systematised concept in a structured way. This is to ensure that the final measurement is sufficiently indicative of what it is supposed to be indicative of. 

For years, scholars have pointed out that computer scientists mostly fail to provide clear concept definitions and evidence-based construct validation~\citep{jacobs-2021-measurement, raji2021ai, liu-etal-2024-ecbd}. A recent study delivers empirical support for this critique through a systematic review of 445 large language model (LLM) benchmarks. More than half of those were found to be designed on the basis of contested concept definitions or no definition at all, and a bit less than half of these benchmarks were published without any reported construct validation in the form of a justification rationale or empirical evidence~\citep{bean2025measuring}. The systematic lack of (evidence for) construct validity renders many benchmarks largely useless, misleading, and guilty of overselling claims. This should alarm us, especially considering the significance of benchmarks in our current practices and debates surrounding AI as a subject of research, a business product, a canvas for projecting our most and least desired futuristic scenarios, and so on.

\subsection{Bias and Misrepresentation}
\label{bias}
The next issue is, in a sense, a subset of the problem of \textit{what} a measurement instrument, such as a benchmark, \textit{is actually} measuring. In the context of AI benchmarking, the issue of data bias can be understood as a form of miscalibration of the instrument towards a data distribution that is not representative of the distribution the instrument was (presumably) intended to be calibrated to. \citet{kraft2025socialbiaspopularquestionanswering} analysed 20 question-answering benchmark datasets and found that several were over-representative of questions related to, e.g., male individuals and Western locations. Even dedicated \textit{bias benchmarks} are not immune to bias~\citep{powere2024statistical,demchak2024assessing}. Just as any dataset, benchmark datasets are \textit{situated}~\citep{raji2021ai} and are likely to reflect the priorities, concerns, realities, conceptualisations, etc. of their creators. In \citet{kraft2025socialbiaspopularquestionanswering}, popular benchmarks were predominantly created by Western, elite institutions. A benchmark that is marketed as measuring the ability to correctly reproduce facts about the world, but mostly contains questions concerning facts relevant to the Western world, is effectively misleading. What is more is that, when applied, such benchmarks effectively reward biased model outputs and feed into algorithmic bias in AI systems~\citep{bowman2021what}.\footnote{\citet{uzunoglu2025flaw} experimentally show that imbalanced representation of `subdomains' within a benchmark dataset, paired with a metric that is based on averaging, yields measurements that obscure potential weaknesses on lesser represented domains. If we conceive of, for instance, different demography-related data points as subdomains, this phenomenon can be expected to generalise also to the type of dataset bias discussed here.} 

This worry should not be understood as only epistemic---in the sense that biases are technical distortions affecting individual measurements---as the score is neither merely academic nor containable to the context in which it was produced. Rather, once adopted, such scores exert an influence that hardens in various ways and domains. Accordingly, these methodological shortcomings cannot be assessed (nor addressed) independently of this social context (nor without attending to various harms discussed in Section~\ref{faces}). 

\subsection{Naturalisation of `Ground Truths'}
\label{naturalisation}

Benchmarks are used by different stakeholders to classify `good' from `bad' or `safe' from `unsafe' models, which shapes their choice of which model to develop further or deploy in real-world applications. Through such choices, the community has witnessed many benchmarks become \textit{de facto} standards in their practice and---as with any standard embedded in practice---this has a ``material force in the world''~\citep[][p. 39]{bowker1999sorting}. The European Union (EU) AI Act, for instance, requires the benchmarking of ``accuracy, robustness and cybersecurity'' in high-risk systems (Regulation (EU)~\citeyear{aiact}/1689), which determines if a model will be allowed to be deployed on the market. Another example of how benchmark scores become natural `ground truths' about a phenomenon is the widely adopted use of the \textit{RealToxicityPrompts} benchmark, which utilises the (soon-to-be terminated) Perspective API for the automated output scoring of hateful and toxic language.\footnote{\url{https://perspectiveapi.com/#/home} (access date: May 6, 2026)} This has become the presumed and accepted standard for toxicity classification and---despite its many conceptual and implementation limitations and inherent biases~\citep{hartmann2026byebyeperspectiveapi}---is used in responsible AI assessments as part of larger LLM evaluation suites~\citep{bommasani2023}.

As such, AI benchmarks give rise to competitive dynamics, with the `winners', i.e., the labs that manage to beat the current state-of-the-art (SOTA), receiving praise, recognition, downloads, citations, and public and institutional trust. Hence, there is a strong pull towards benchmarks, motivating the community to design systems that will be able to compete, which, consequently, structure the community’s innovation efforts~\citep{orr2024aisport}. In short, benchmarks determine the type of systems that are built and which truths they are calibrated towards. Once benchmarks acquire this status, their flaws do not merely mislead specialists but also enter the broader public and commercial imaginary of AI. Weak measurements, biased representations, and \textit{de facto} standards thus serve to fuel myths about what AI systems are and what they are becoming.

\subsection{Fuelling the Myth}
\label{myth}

The global `AI race' motivates well-funded labs to outdo their competitors, and the pace, scale, and diversity of improvements have led to a state where established benchmarks become too easily solved or simply outdated. In response, benchmarks are created to pose new challenges to the field or to prove `human-like' capabilities. The authors of \textit{Humanity's Last Exam} (HLE)~\citep{center2026benchmark}, for instance, claim that ``[h]igh accuracy [...] would demonstrate expert-level performance on closed-ended, verifiable questions and cutting-edge scientific knowledge''. 

Above, we argued that many established benchmarks are highly misleading as they do not measure what they claim to measure, while systematically concealing concerns related to social bias. This calls for a re-assessment of several beliefs about AI that have been circulating in different areas of society and actively promoted by Big Tech. We argue that claims about human-like or close to human-like `reasoning' or `knowledgeability' levels should be met with scepticism, because those profiting most from these anthropomorphising framings are those who claim to be most familiar with the state of research. Hence, we deem it plausible and necessary to assume that there is another reason to continue overselling benchmark results and that this may be of economic and political nature, as opposed to the mere pursuit of knowledge about AI technology's real capabilities. As it stands, many highly visible benchmarks achieve nothing more than: (1) mislead the public into thinking that AI systems become more and more `human-like', and (2) facilitate gatekeeping that excludes users, policymakers, journalists, and other relevant stakeholders by making it appear as though AI systems are beyond comprehension. As such, AI is also made to appear as though those stakeholders cannot truly contribute to shaping it. When scores enable not only the ability to define what `counts' but, in the process, confer reputational and commercial advantage, an almost cliché temptation to game the established system emerges. 

\subsection{Gaming of the System}
\label{gaming}

As guides for selection and deployment, AI benchmarks become an important lever for marketing. With regards to Meta's Llama~4 model family which was released in 2025,\footnote{\url{https://ai.meta.com/blog/llama-4-multimodal-intelligence/} (access date: May 6, 2026)} the company's former Chief AI scientist, Yann LeCun, in fact, stated in an interview ``that the 'results were fudged a little bit,' and the team used different models for different benchmarks to give better results.''\footnote{\url{https://arstechnica.com/ai/2026/01/computer-scientist-yann-lecun-intelligence-really-is-about-learning/}, (access date: May 6, 2026)} When OpenAI announced the release of the o3-model, at the end of 2024, it advertised its capabilities as groundbreaking, among other things, as measured by a particularly difficult mathematics benchmark called \textit{FrontierMath}.\footnote{\url{https://techcrunch.com/2024/12/20/openai-announces-new-o3-model/} (access date: August 15, 2026)} This benchmark had been created by the company EpochAI who contracted expert mathematicians to author difficult mathematics problems with the goal of testing a new generation of `reasoning models'. OpenAI had not only funded the creation of FrontierMath, but both companies had also agreed to remain secretive of their partnership until the release of the o3-model.\footnote{\url{https://techcrunch.com/2025/01/19/ai-benchmarking-organization-criticized-for-waiting-to-disclose-funding-from-openai/} (access date: May 6, 2026)} Moreover, they had only verbally agreed that OpenAI would not train their newest model on the benchmark. 

It is crucial that a model is not directly trained on any of the examples that occur in a benchmark that it is later evaluated by. This phenomenon---called \textit{data contamination}---must be avoided to ensure that the evaluation score is a true measure of model performance and not only a result of memorisation~\cite[cf.][]{schaeffer2025causally, sainz2023nlp}. The `gaming' of the AI benchmarking system can also happen in more implicit ways. It is, for instance, assumed that to reach high scores on the coding benchmark \textit{SWE-Bench}~\citep{jimenez2024swebench}, developers ``craft [...] approaches that are too neatly tailored to the specifics of the benchmark'',\footnote{\url{https://www.technologyreview.com/2025/05/08/1116192/how-to-build-a-better-ai-benchmark/} (access date: May 6, 2026)} meaning that the training examples are selected to closely resemble those presented during testing, which yields effects similar to memorisation.\footnote{In the meantime, OpenAI has published its own analyses illustrating data contamination of flagship LLMs regarding this benchmark; \url{https://openai.com/index/why-we-no-longer-evaluate-swe-bench-verified/} (access date: May 6, 2026)}

The deal between OpenAI and EpochAI undermines the necessary conditions for meaningful and trustworthy benchmarking practices. Indeed, such behaviour invalidates the very idea of AI evaluation. While it is hard, if not impossible, to prove widespread intentional misconduct, cases like the FrontierMath scandal, as well as the admitted manipulation of benchmark results to exaggerate Llama 4's capabilities, are clear warning signals: Benchmarks have become a tool of power which resourceful institutions are not hesitating to use towards their own ends. 

Together, these five interrelated concerns highlight why the trouble with benchmarks cannot be reduced to a list of technical defects. Benchmarks transform uncertain and situated measurement practices into public claims about quality, safety, and progress. Their flaws matter because benchmark results circulate far beyond the conditions under which they are produced. Growing recognition of the these limitations does not appear to have diminished the importance of benchmarks and, in particular, the intensity with which they are used to promote LLMs. Benchmarks continue to be used because they are an important structural element of the AI development culture and, so far, there is a lack of a better alternative. This suggests that their authority cannot be explained by measurement quality alone, and that the problems---highlighted in  Sections~\ref{naturalisation},~\ref{myth} and~\ref{gaming}---extend beyond technical fixes and pertain rather to the choices and actions of those seeking to market or steer community efforts towards their own goals. Closer scrutiny of the nature of these operations and development practices---of the kinds of relations and power structures they give rise to, who is being pushed out as a result and at what cost---is, therefore, warranted in order to appreciate various harms, we argue, that follow from the just described dynamics. The harms outlined below are, therefore, not offered as a second, independent, set of objections, but rather specify what is normatively at stake as complicating factors for tackling the issues above as well as demanding redress in their own right (as \textit{injustices}).   

\section{Benchmarking and the Faces of Oppression}
\label{faces}

Due to how cases like FrontierMath or the beautified evaluation of Llama 4 leverage an unfair advantage over other competitors, most would likely agree that some wrong has been committed. Yet, a claim that this also amounts to an `injustice' or is `oppressive' would likely be contentious. Nevertheless, we argue that features of existing benchmarking practices are not merely regrettable and warrant remedial efforts but also perpetuate forms of oppression. 

For~\citet{young-1990-justice}, injustices denote a specific kind of pervasive, systemic, and instituted wrong that occurs when social processes routinely privilege some groups while disadvantaging others. She explains that injustice primarily refers to two forms of disabling constraints: \textit{oppression} and \textit{domination}. Oppression concerns the myriad ways in which socially instituted mechanisms undermine or limit self-development and expression, while domination is reserved by Young to capture instances of political exclusion that bar groups from decision-making procedures or formal institutional control mechanisms, thereby undermining self-determination. Accordingly, social justice is about promoting and securing both self-development and self-determination \citep[see][p. 702]{Naudts_2024}. 

While we believe that an argument could be made that benchmarking practices can be dominating in Young’s sense, defending this would require detailing and unpacking the many still unfolding instances of political influence and lobbying perpetrated by the AI industry, and propped up by benchmarks, which space forbids doing meaningfully here. As such, in this paper, we will focus only on the notion of oppression. Famously,~\citet{young-1990-justice} identifies ``five faces of oppression''---\textit{exploitation}, \textit{marginalisation}, \textit{powerlessness}, \textit{cultural imperialism}, and \textit{violence}---the first four of which, we argue, arise in the case of AI benchmarking.\footnote{It is worth briefly noting that benchmarks do also, at least indirectly, serve a function in perpetuating salient forms of violence. \citet{Naudts_2024} notes that ``[o]nce data-driven systems have normalized a particular set of norms, values and beliefs in a specific setting, deviancy can be identified and made subject to interference'' (p. 6) and AI benchmarks act as a shared blueprint and certifier for detectors of conformism and deviancy. Benchmarks thus indirectly shape and grant legitimacy to violence-promoting systems.} Importantly, the faces are not necessarily exclusive, as a single benchmarking practice (or facet thereof) may instantiate several, which nevertheless capture distinctive kinds of wrongs that ought to inform any attempted reforms. Exploitation concerns the transfer and capture of labour and benefits; marginalisation, exclusion from recognised participation; powerlessness, participation without authority over its terms; and cultural imperialism, the universalisation of one evaluative standpoint.

We are not, of course, the first to apply Young’s notions in the context of AI and algorithmic decision-making. Yet, the existing literature has primarily focused on two aspects, namely, data- and model-specific practices and features (e.g., giving rise to concern around bias, transparency, opacity, explainability, consent, etc.), on the one hand, and the usages those technologies are put to (e.g., negatively impacting individual lives and livelihoods), on the other \citep[cf.][]{Browne2023AIStructuralInjustice,Herzog2021AlgorithmicBias,HerzogBranford2025RelationalEthics,Naudts_2024}. Our focus on AI benchmarking, however, highlights underexplored elements of the broader AI ecosystem. AI development proceeds iteratively through conceptualisation, data acquisition and preparation, model training and evaluation, and deployment, usage, and monitoring, with feedback loops from model evaluation informing which data are added to the model training and how the model design and training are further adjusted (\citet{bordes2025evalfactsheetsstructuredframework} call this development-focused evaluation). As the SWE-Bench case demonstrates, to outperform earlier models, researchers and engineers repeatedly adjust systems through result-guided experimentation. Evaluation results directly influence which models are put to real-world use and serve the AI research community as indicators of progress,  constituting an enclosing steering mechanism that, we will argue, is normatively consequential. In particular, we consider three instances of related harms: (1) ethical harms perpetuated against those involved in AI development due to inequitable competition; (2) ethical and epistemic harms stemming from AI systems with inferior performance, robustness and with problematic biases, all of which are concealed by invalid or miscalibrated benchmarks; and (3) epistemic harms that pertain to the scientific and technological field as a whole due to flawed and misguiding indicators of progress.

Note that, in presenting the below harms as widespread structural concerns, we do not claim that all researchers and developers who create and apply benchmarks are blame-worthily complicit. These issues result from collective activities and standards formed over time through many small, local, and often well-intentioned decisions (see Section~\ref{structural}). Still, we distinguish between `powerful players'---well-funded, highly visible, and influential industry labs---and less influential researcher and developer groups. The former are not accused of having imposed this system on everyone, but of leveraging and shaping it. They have a greater capacity to ameliorate the issues identified and, therefore, bear a greater share of the responsibility for doing so. Still, collective action will certainly be needed to motivate this, and so responsibility is also dispersed across the field.  

\subsection{Exploitation}
\label{exploit}
Talk of exploitation in AI is likely to evoke images of precarious, harmful, low-paid data labour, asymmetrical and abusive power relations, and large-scale appropriation of data and intellectual property \citep[cf.][]{Hao2025EmpireOfAI,Crawford2021AtlasOfAI,MuldoonGrahamCant2024FeedingTheMachine}. While these deplorable practices are also evident in AI evaluation datasets, affecting the stakeholder group of data authors and annotators (including those whose data were extracted without their knowledge and consent), they do not exhaust the kinds of exploitation evident in this context. According to Young, exploitation refers to structural relations and ``social processes that bring about a transfer of energies from one group to another to produce unequal distributions, and in the way in which social institutions enable a few to accumulate while they constrain many more''~(\citeyear[][p. 53]{young-1990-justice}). Exploitation is thus a patterned conversion of the burdens, risks, and productive energies borne by (or extracted from) some actors into benefits that accrue elsewhere. 

AI benchmarks are coordination devices that organise labour, attention, prestige, and investment among involved stakeholder groups, i.e., researchers and engineers~\citep{raji2021ai, orr2024aisport}. They create an environment in which it is rational, and often professionally necessary, for researchers to direct their labour toward improving a narrow set of public scores. That labour may not, in itself, be objectionable. What is objectionable is that the terms on which it is solicited, conducted, and rewarded are routinely set by actors positioned to convert benchmark outcomes into proprietary advantage, reputational capital, and agenda-setting authority that primarily serves their own narrow interests, all the while giving the impression that such efforts are necessary for advancing the field as a whole \citep[cf.][]{Hao2025EmpireOfAI}.

Those who define a benchmark’s make-up---especially when they can also significantly influence community uptake---can, thereby, exert control over the purposes to which communal labour is put (i.e., the research trajectory), the criteria under which that labour is valued (i.e., metrics of success), and the conversion of resulting gains into institutional and economic advantages \citep[cf.][]{Ott2022MappingGlobalDynamics,ThomasUminsky2022RelianceOnMetrics}. The extent to which benchmarks come to gain scientific authority often obscures this. Put differently, this exploitative power is often exercised through ostensibly technical design decisions that present as merely methodological, but are actually highly normative.\footnote{\citet{marin2026aimodelaccurateenough} emphasise that several \textit{techno-normative choices} are made when defining and implementing evaluation measures, and the mere choice of metric (e.g., accuracy versus precision or recall) can cause rare cases of a disease to be ignored or social media statements to be incorrectly flagged as problematic.}

As a result, such decisions structure the field’s incentives in precisely the way Young’s analysis anticipates: they render certain forms of effort intelligible as `progress' and others as peripheral, thereby channelling labour toward aims that disproportionately serve already-powerful institutions. One might object that this overstates the agency of benchmark creators. After all, benchmarks are in principle open. However, while less resourced academic labs may produce incremental gains that become part of the field’s shared `improvement curve', the capacity to consolidate these gains through scale, proprietary deployment, marketing, or integration into widely used services largely remains concentrated in well-capitalised institutions. Exploitation, on this reading, consists in the field’s collective energies being channelled through evaluative rules that many must comply with but few can meaningfully contest, thereby entrenching dependence on standards others control, and limiting contestation over what and who AI development is for.

\subsection{Marginalisation}
\label{marginal}
While exploitation concerns how energies are transferred and benefits captured, marginalisation concerns who is kept out of the practices and discussions in which purposes are set, criteria are stabilised, and `progress' is publicly ratified.

Several popular LLM benchmarks are demographically and geographically biased (biases towards Western, Christian, and male subjects have been identified) and their reporting is opaque, e.g., regarding the identities of annotators~\citep{kraft2025socialbiaspopularquestionanswering}. All the while, the most influential benchmarks are created by a few elite institutions~\citep[see also][]{koch2021reduced}. Benchmarks calibrated to these select perspectives and interests ultimately incentivise the development of models that are biased accordingly. This is because LLM training is always a cost-benefit calculation and data are expensive to obtain, annotate, filter, and process. %So, labs only collect the data needed to optimise their model such that it succeeds in solving whatever problem configuration the community is endorsing. 
From such a vantage (objectionable as it is), collecting data representing knowledge from Eastern or Southern regions is disincentivised, given that it would not enhance performance as measured by a benchmark that is calibrated to Western knowledge. Community endorsement of biased benchmarks, thus, solidifies the use and reuse of biased training corpora and, finally, the development of biased models. In these cases, benchmarks are complicit in widely discussed forms of marginalisation, e.g., when AI systems discriminate against particular groups who are, as a result, systematically denied economic opportunities~\citep[cf.][]{Bommasani_2022_algorithmicmonoculture, Jain_2024_algorithmicpluralism, Naudts_2024} or harmfully misrepresented~\citep{blodgett2020language}. Biased benchmarks certify the `accuracy' of systems offering discriminatory output and form part of the impetus to adopt such systems that go on to cause harms of this kind. 

Young explains that ``[m]arginals are people the system of labour cannot or will not use''~(\citeyear[1990][p. 53]{young-1990-justice}), which accurately describes the situation of researchers and engineers who actively seek to participate in the community efforts to innovate and understand AI but are either entirely precluded or systematically pushed to the periphery. In contemporary AI research contexts, this marginalisation manifests not only as material deprivation but through the withholding of certain kinds of standing, e.g., by having one's work overlooked. Consequently, a ``whole category of people is expelled from useful participation'' (Ibid), who are  ``rendered invisible, voiceless, unrecognised and isolated'' \citep[][p. 705]{Naudts_2024}.  Rather than direct their labour more freely toward ends they deem morally valuable or scientifically promising, these researchers are (at least indirectly) compelled either to chase established benchmarks or to build `competitors' that nevertheless respond to those benchmarks and so cannot fully shed their terms. 

Also marginalised by benchmark-centric research are those who lack the resources to engage in the benchmark race at all. As cutting-edge benchmarks increasingly presuppose access to large-scale compute, proprietary datasets, and extensive engineering labour \citep{Strubell2019EnergyPolicy,Bender2021StochasticParrots}, benchmark leaderboards often track access to resources as much as (or more than) methodological ingenuity, scientific value, or social utility \citep[cf.][]{orr2024aisport,Ott2022MappingGlobalDynamics}. The consequence is that entire research communities (e.g., those based in less-resourced institutions, regions, or disciplines) are effectively marginalised. It may be contested that this is unavoidable, as some kinds of research are simply costly and the price of advancing the field. We reject this as a self-fulfilling prophecy resulting from the current benchmark system, which perpetuates a narrow conception of what constitutes advancement. 

There are AI researchers innovating the field whose work is not captured by sitting atop prominent leaderboards. For instance, most popular LLM benchmarks measure English-specific performance only~\citep[cf.][]{kraft2025socialbiaspopularquestionanswering}. While the natural language processing (NLP) community has seen an increase in efforts to build systems attuned to `low-resource' languages, yet visibility and praise remain rather limited. Such dedicated benchmarks do not play a significant role in the public and more established benchmarks that do cover `lower-resource' languages are usually those designed to measure multilingual capabilities as an aggregate concept and are based on translations or adaptations from originally English benchmarks, which come with their own quality issues~\citep[cf.][]{umutlu-etal-2025-evaluating}. If model performance on these less represented languages was considered central to AI innovation, respective benchmarks would receive more attention, directing collective efforts accordingly. Not only would this benefit these language communities but also those who have been researching and developing in these areas all along. 
Our point is not primarily about such work getting its due in terms of visibility, but rather that the benchmark regime structures what counts as intelligible achievement in the first place. Certain kinds of work (e.g., locally grounded evaluation, qualitative assessments of harms, low-resource language work, or alternative task framings) are institutionally disincentivised and rendered epistemically secondary, signalling that such work is less important for the field. What is needed are expanded possibilities for marginalised researchers and labs to do the work they believe will enhance the field without being relegated as a result.

It is tempting to treat marginalisation here as primarily representational. Diversification in the current benchmarking arena is largely lip-service as self-determination (i.e., the ability to set the values and directions of the field) remains severely limited~\citep[cf.][]{birhane2022power}. Critical work on data colonialism and decolonial approaches to AI has repeatedly illustrated how `inclusion' can function as a continuation of colonial patterns of appropriation liable to being repurposed for institutional or commercial ends~\citep{couldry2019datacolonialism,MohamedPngIsaac2020DecolonialAI, Ricaurte2019DataEpistemologies, Ricaurte2022EthicsMajorityWorld}. 
The worry is that, without restructuring existing benchmarking practices and their influence on the field, inclusion of marginalised communities may be pursued primarily to improve dominant metrics, expand market reach, or strengthen claims of generality. Accordingly, `difference' is incorporated only insofar as it can be translated into the prevailing evaluative norms, while the communities' own standards of success (e.g., what counts as a good model, harmful output, or legitimate use case) remain excluded.

\subsection{Powerlessness}
\label{powerless}
If marginalisation concerns exclusion from meaningful participation, powerlessness---as just hinted at---concerns participation without authority. It is about whether an individual's judgement counts as judgement that will be heeded to improve the conditions of their life, whether they can speak in a given setting without being marked as out of place, and whether their participation includes the ability to contest and reshape the terms under which they are evaluated. As such, an individual can be quite involved in the practice---and, as in the case of AI researchers, be recognised as skilled and competent---yet still occupy a subordinate role in entrenched practices that limit their action and inhibit meaningful authority over how and to what ends they labour. 

A fine-grained appreciation of the powerless entailed in AI benchmarking requires disentangling some elements of Young’s account, since her analysis is fundamentally concerned with `class', tracking a division between `professionals'---who exercise authority and discretion, and who are granted recognition---and `non-professionals'---who are supervised, expected to follow orders, and routinely disrespected. Further, Young extends such workplace practices into society more broadly where such standing or lack thereof continues, through feedback loops, to exert influence and shape social expectations and interactions. Accordingly, the powerless are diminished across the board, severely inhibited from developing skills, exercising creativity or judgment, expressing themselves in a manner that is heard, or sharing their experiences (and the positional knowledge entailed therein); in short, means to improve their situation are effectively immobilised. 

The positions of data workers and other subordinated labourers in the AI supply chain closely resemble the non-professional status Young has in mind~\citep[cf.][]{Naudts_2024}. One might argue that data annotators do own a certain level of power in the sense that they can shape which data are considered `right' or `wrong', `desired' or `undesired'. However, due to their material dependency paired with strict annotation criteria and tightly regulated and supervised work environments, their annotations are actually more likely to reflect those of the clients (i.e., companies like OpenAI)~\citep{miceli2022dataproduction}. 

Focusing on benchmarking as a research-governance practice, two further constituencies matter that complicate Young’s professional/non-professional distinction. First, there are \textit{users} whose powerlessness flows from the extent to which they are drawn into a relation of epistemic deference concerning AI models. Benchmarks are the dominant paradigm to diagnose accuracy, robustness, and security (as reflected in the EU AI Act, Article 15(1)), affecting operator and regulator decisions as well as informing the wider public of the current state of AI (e.g., mediated by journalism). As discussed in Section~\ref{myth}, evaluation scores can thus fuel certain (mythical) conceptions of AI irrespective of numerous concerns related to bias or invalidity. This evaluation infrastructure---reified by legislation---renders the general public powerless. That is, the in-transparency of the practice deprives the public of its means to form a comprehensive position of its own. Benchmark scores and leaderboard rankings function as ready-made `evidence' for what is `best', which invites users to treat the relevant metric as reliable and the use of leading models as responsible. Stretching Young’s language of `class'---even though users may be `professionals' who deploy existing models in their own downstream applications---the division becomes between those who are able to set the conditions of evaluation and develop systems able to best meet them, on the one hand, and those who must take this all on face value. This can be conceptualised as a kind of epistemic powerlessness, defined by being governed, in one’s practical decisions, by evaluative machinery one cannot meaningfully contest, audit, or reinterpret.

Second, there are the \textit{researchers} actively designing their own benchmarks, developing models, writing papers, etc. These actors are highly skilled, articulate, and credentialed. They look like `professionals' in Young’s sense, and so---by definition---should be excluded from the powerless. Yet, even though anyone can contribute their own benchmark, not all benchmarks are destined to be considered a widely shared standard. As with any research contribution, relevance to currently shared areas of interest, improvements over existing approaches, and visibility within the community are preconditions. Once performance on a specific benchmark becomes the currency of success, many researchers are pressed into a role that looks less like collective authorship and more like a form of deference and adherence to tasks confined to mere optimisation, comparison, and reporting within the space delineated by the given benchmark paradigm. Orr and Kang’s diagnosis of benchmarking as a kind of `sport' is helpful, as sport is a competitive practice governed by rule-makers, officiating procedures, and legitimacy-conferring institutions that decide what counts as a win~\citep{orr2024aisport}. This is not to say that professional athletes are powerless in an absolute sense, and neither are AI researchers and engineers. But current benchmarking practices, and in particular their capture through well-funded institutions with political and economic agendas, set significant constraints on the assertion of power. Researchers can act, but only within terms they did not set and which may draw them into activities they may object to. That is because refusing the benchmark race and insisting on alternative criteria of success is not only likely to carry significant professional costs, but their activities may indeed be in some sense unintelligible within the dominant economy of recognition perpetuated by benchmarking and, therefore, likely to be dismissed. This already suggests that the trouble resides in the broader system. 

The gravitational force of benchmarks invites myopia, gaming, and the displacement of other substantive goals~\citep{ThomasUminsky2022RelianceOnMetrics}. Even researchers who are personally committed to aims like social utility or responsiveness to harms are likely to find themselves having to translate those commitments into benchmark legibility, or to treat them as secondary rather than as success conditions in their own right. The result is a domain in which many `professionals' are still, in a decisive sense, highly limited authors of the norms that govern their professional lives. 

\subsection{Cultural Imperialism}
\label{imperialism}
For Young, cultural imperialism refers to the social processes under which a dominant group can universalise and establish their experiences and culture as the norm. Conversely, the particular perspectives and lived experiences of less privileged groups become obscured, stereotyped or marked out as `other'. The harm is that dominant culture secures the status of common sense, appearing `natural' or authoritative, thereby subsuming its way of seeing (or sense-making) as standing in for reality as such, while alternatives appear partial, parochial, or deficient in some way. 

\citet{Naudts_2024} transposes this point into the digital ecosystem by arguing that data-driven systems increasingly function as modern ``means of interpretation and communication'' in that they order, classify, and represent the world in ways that can naturalise particular values as technical facts. AI benchmarking practices exemplify this dynamic. As evaluation encircles the whole AI development and deployment pipeline, it directly shapes future model optimisations. Benchmarks are, as such, both complicit in, and even a source of, those discriminatory system behaviours that ``other'' (groups of) individuals detailed by~\citet{Naudts_2024}. They also certify automation systems that are then utilised at scale, ``ordering of the world'' according to dominant views. The worry, thus, is that the current modes of benchmarking have become a dominant---as well as \textit{dominating}---means of interpretation, whereby a particular and exclusionary evaluative culture is normalised and wrongly perceived as neutral, functioning as a shared standard of `right' and `wrong', `desirable' and `undesirable' model behaviour at scale (see Section~\ref{naturalisation}). Indeed, their authority depends in part on obscuring that particularity.

As \citet{hartmann2026byebyeperspectiveapi} highlight through the example of Perspective API, whole communities of scientific practice came to base large bodies of research on one specific, problematic operationalisation of toxicity. Benchmarks serve an illusion of objectivity, formalising the world into binaries, when---upon closer look---each binary is a reductive formalisation that involves actively choosing which error margins to accept~\citep{marin2026aimodelaccurateenough}. \citet{raji2021ai} illustrate that performance on particular benchmark suites is repeatedly made to stand in for broader claims about flexible or general AI capacities, universalising what is inherently a local, value-laden evaluation and transforming a limited utility into an authority. Benchmarks thereby set the interpretive regime through which activity in the field gains meaning. 

This both accentuates the earlier noted difficulties of `powerlessness' while complicating any would-be remedial efforts concerning `exclusion' and `marginalisation' by restricting the conditions of participation in a way that obfuscates that this is in fact a restriction (albeit one that has been internalised by those whose interests this arrangement serves). As a result, benchmarks operate as gatekeeping mechanisms that constrain the scope of self-development and self-determination, compelling actors in the field to orient their own practices accordingly. Due to this universalising capacity, even attempts to break with establishment (e.g., community-driven benchmarks for low-resource languages, local safety concerns, or domain-specific harms) are more likely to have their value recognised when they assimilate to and render themselves intelligible in the benchmark system. In other words, to show that they outperform on a recognised measure. Important research that cannot `travel' in this way (e.g., participatory assessments of model impacts) may struggle to gain traction or be co-opted to legitimate established benchmarks \citep[][]{birhane2022power,Sloane2022ParticipationNotDesignFix}. 

Benchmarks privilege certain research interests and trajectories while marginalising others. However, the problem goes beyond unfair market capture; it negatively impacts the science as a whole by pre-emptively narrowing what can appear as knowledge, critique, or innovation. This, therefore, adds to those issues related to bias (Section~\ref{bias}) and lacking construct validity (Section~\ref{validity}) which already underscore how epistemically fraught benchmarks are and significantly challenge their wide acceptance as meaningful indicators of scientific progress. The corporate capture of AI evaluation under the growing global politico-economic appetite for AI breakthroughs amplifies this issue, eroding an epistemic practice that decreasingly provides meaningful epistemic insight and increasingly serves the rationalisation of dominant norms and profitable myths. The harm done here does not only affect researchers and engineers, but also operators and decision-makers, journalists, and consumers.

\section{Structural Injustice and the Science of AI}
\label{structural}
The four faces of oppression diagnosed in AI benchmarking are different manifestations of the same underlying structures of power we will now investigate through the notion of \textit{structural injustice} proposed by~\citet{young-2010-responsibility} and further developed by~\citet{McKeown_2021}. This structural view reveals how these forms of oppression yield a further instance of harm affecting the epistemic fabric of AI research itself. 

\subsection{An \textit{Avoidable} Structural Injustice}
The instances of oppression laid out in Section~\ref{faces} are structural in the sense that they arise from a culture of evaluation established over time~\citep{orr2024aisport} that is carried and upheld by collective practices. Put differently, they are not merely harms that occur \textit{within} benchmarking but are made possible by the position benchmarking now occupies in the AI ecosystem as a relatively stable arrangement of norms, incentives, resources, institutions, expectations, and interpretive schemes that direct action. We should not, therefore, focus on these harms \textit{solely} as following from the actions of discrete individuals---even though it is typically individuals who often must shoulder their direct costs \citep[cf.][]{Kasirzadeh2022AlgorithmicFairness}---and so seek out or `trace' \cite{Browne2023AIStructuralInjustice} the `wrongdoer', e.g., the team lead who excluded a community driven project, or the business who failed to recognise their lack of diversity, or the company who cheats to get ahead on a benchmark. While in many cases it will be possible to identify some kind of wrongdoing and assign a portion of responsibility for the injustice in question, this does not \textit{fully} capture what is going on in benchmarking. Rather, as in the case of what  \citet{McKeown_2021} calls \textit{pure} structural injustice, these also follow from people systematically acting in morally acceptable ways with no ill intent to others or intention to exploit a situation for their gain, and so stem from largely well-intentioned, normalised, and seemingly unobjectionable actions of individuals exhibiting network effects~\citep{young-2010-responsibility}. Concretely, it can be assumed that most researchers and developers are genuinely interested in developing the best possible measures for quality, safety, and progress. This is reflected in increasing collective efforts to improve AI evaluations (see Section~\ref{implications}).

However, \citet{McKeown_2021} nuances Young's notion of structural injustice, differentiating two further kinds that better track the aspect of `power' (pp. 4-5).
\textit{Avoidable} structural injustice is where there are powerful agents with the capacity to change unjust structures, but they fail to do so. For example, rich states have the capacity to end homelessness, but they do not. So where Young understood homelessness as a structural injustice simpliciter, McKeown argues that it is avoidable. \textit{Deliberate} structural injustice is where agents are deliberately perpetuating unjust background conditions for their own gain, and they have the power to change them. For example, in the sweatshops case, multi‐national corporations deliberately lobby governments under the threat of capital flight, in order to maintain the poorest wages and working conditions for garment workers. McKeown concludes, therefore, that when powerful agents deliberately perpetuate structural injustice or have the capacity to change unjust structures, they bear moral responsibility to do so.

It is difficult to categorise AI benchmarking as a whole, since the landscape of benchmarks, what they are built and used for, and how and whether they become established as a collective standard, is quite diverse. Some benchmarks emerge from shared tasks or are published as test sets in AI publications, others are directly created or commissioned by technology companies. Nevertheless, benchmarking has the hallmarks of \textit{pure} structural injustice given that no single actor creates the benchmark race and the structure is reproduced through the ordinary, often well-intentioned activities of an array of stakeholders adhering to established norms. Yet, the relevant harms are no longer unforeseeable, nor are all participants equally positioned. The benchmarks that come out as the most popular are increasingly those created by well-funded research and industry labs~\citep{kraft2025socialbiaspopularquestionanswering, baack2026unsteadymetricsbenchmarkingcultures}. Technology companies, in particular, continue to reap the benefits instead of triggering change in favour of fair and inclusive evaluation. In this sense, we argue, the injustices observed in AI benchmarking fall into the category of \textit{avoidable} structural injustice. The running system was not put into place to deliberately cause harm and, in fact, it is plausible to assume that most stakeholders both intend to avoid harm and seek to advance the field in the best way they know how. Nonetheless, in its current state, influential players hold unequal power in and over this system, and do have a real capacity to alter the current setup, which means that they can choose to either continue to leverage it to their own benefit or change it in favour of others (or mitigate the concerns raised above).

\subsection{How Benchmarking Harms AI Research}
Acknowledging this structural dynamic permits us to step back from specific instances of injustice and assess how the system they constitute may weaken AI research as a whole. Writing in the context of AI decision-making in medicine, \citet{HerzogBranford2025RelationalEthics} flag a concern directly relevant to ours. They note that AI use and reliance may be ``indulging in an inadequate epistemological system that oppresses knowledge production and possession of a particular kind without us having the capability to recognize this'' (p.15). By diminishing physician and patient involvement, a particular conception of what counts as `good' medical practice is being advanced and barriers erected that inhibit the recognition of shortcomings and the means for assessing the value of alternatives. A similar fear underscores the epistemic concerns we have vis-á-vis benchmarking. Specifically, that the current model not only excludes alternative conceptions of what counts as `good' AI development or `progress', but that it is itself structured to reward a distinctive perspective concerning this. It is, therefore, necessary to reflect on whether the prominence of benchmarking has produced a similarly ``inadequate epistemological system''. 

Echoing concerns raised by~\citet{dotson2012cautionarytale}, we argue that the systems of evaluation in AI are calibrated to exclude certain groups from participating in this significant epistemic process. The interests of less dominant researcher, developer, and consumer groups are inadequately represented in the established frameworks, which has two forms of ethically and epistemically harmful consequences: First, it leads to insights about AI progress and appropriateness that are less relevant to their preferences and experiences. Due to the causality between evaluation, development, and deployment, this yields systems unfit to their needs and wants, and which perpetuate bias and misrepresentation. Second, current frameworks prevent objectivity. For many years, philosophers of science have argued that objectivity can only be approached via temporal and local consensus that is best justified through ongoing engagement with diverse, conflicting, situated, partial, and changing understandings of the matters in view~\citep{Longino2002, haraway2016situated}.

To fix the science of AI, the community needs to pursue what~\citet{dotson2014conceptualizing} would categorise as \textit{third-order change}: AI research is trapped in a closed cycle, in which those who determine the mode of evaluation are those who build the systems which are to be evaluated, and those who benefit from their uptake~\citep{kraft2025knowledge}. To overcome epistemic oppression upheld by current practices, we must start with the epistemic tools we utilise to judge and tune them. Escaping the evaluation crisis is not only a matter of building more reliable benchmarks. It is also a matter of questioning the concept altogether and the currently accepted norms related to the \textit{how}, \textit{who}, and \textit{for whom} it is conducted. It must, therefore, involve transforming the conditions under which evaluation becomes authoritative, the purposes it serves, the agents empowered to define it, and the institutions that convert their results into material and symbolic power.

\section{Implications for the Future of AI Research}
\label{implications}

We have argued that AI benchmarking is not merely a technical practice for measuring progress, but a socially powerful evaluative infrastructure that shapes labour, recognition, authority, and the direction of AI research that helps perpetuate instances of four out of Young's five faces of oppression---exploitation, marginalisation, powerlessness, and cultural imperialism---within the AI ecosystem. Moreover, we have argued that these arise as not only a form of \textit{structural injustice} but an \textit{avoidable} form which is currently epistemically and ethically undermining AI research. But how do we fix an ``inadequate epistemological system''? How can we possibly transcend our established frameworks and alter our known ways of doing AI research from within? 

Efforts to establish an ``evaluation science for AI'' have been surging, calling for deeper engagement with the statistical foundations of testing, more mathematically rigorous measurement instrument creation, or more transparent evaluation documentation and sharing of results \citep[cf.][or EvalEvalCoalition\footnote{\url{https://evalevalai.com/research/2025/07/13/eval-science-kickoff/} (access date: May 17, 2026)}]{truong2026measurement}.
We clearly welcome this collective push towards more valid, reliable, and transparent benchmarking. However, we also warn against circling around individual symptoms (especially in isolation) and technosolutionism. We argue that we must not focus too narrowly on an `inside-only' view on issues and, instead, encourage to zoom out, to conduct power-critical inquiry, and understand the structural entanglements of evaluation. Indeed, it is precisely the need to do so that is highlighted by the accumulative instances of oppression detailed in Section~\ref{faces} and their relation to the issues noted in Section~\ref{scene}. Again, revising the epistemic system in place calls for \textit{third-order change}, which ``involves developing the capacity to recognize and alter elements of operative, instituted social imaginaries that inform and preserve organizational schemata''~\citep[][p. 119]{dotson2014conceptualizing}. However, because benchmarks constitute the very epistemic schema with which the community is used to judge its practice, assessing and revising the concept as a whole is challenging, in particular, for those who are situated within this community~\citep[see also][]{kraft2025knowledge}. Therefore, `outside' perspectives are likely to prove invaluable~\citep[cf.][]{Branford2025GenerativeAI}.  
 
Benchmarking is inevitably a reductive mode of analysis and, in important respects, may well remain \textit{per se} irreparably problematic and prone to capture. Yet, this is not an isolated problem, limited to AI, but one long recognised in the philosophy of science, which has also suggested that we can nevertheless structure our practices to best mitigate these deficiencies by accepting the situatedness of knowledge claims and being explicit about their purported utility (understood as a value-laden claim). Specifically, this leads us to advocate for AI research to endeavour towards what \citet{Kitcher2001ScienceTruthDemocracy,Kitcher2011ScienceDemocraticSociety} terms a ``well-ordered science'', which is intended to ``serve the collective good'' (2001, p. xiii) by identifying relevant ends and means through the best available approximation of an ``ideal conversation, embodying all human points of view, under conditions of mutual engagement'' (2011, p.106). Through this it might aspire to ``answer the \textit{right} questions in the \textit{right} ways, where value judgements and methodological issues are inextricably intertwined in determining what is \textit{right}'' \citep[][p. 981]{Cartwright2006WellOrderedScience}. Accordingly, the path towards better AI evaluation requires (1) a wider, participatory ecosystem of imagination, debate, and audit, while at the same time (2) accepting and embracing the locality and temporality of what is considered `right' and `wrong', `desirable' and `undesirable'. 
 
Regarding (1), the AI community is already trialling more collective forms of evaluation, namely \textit{red-teaming}, i.e. collective adversarial testing of `model safety'~\citep{ganguli2022redteaming}, and open voting platforms like Chatbot Arena~\citep{chiang2024chatbotarena}, where anyone can rate models against each other according to one's preferred outputs. However, \citet{parth2024democratic} raise the concern that those approaches are falsely marketed as `democratic', and that powerful players are harvesting free labour under this guise. The crowd-sourced performance judgements and preference feedback (on reductive pre-defined dimensions and in strictly standardised formats) are usually absorbed by technology companies towards the improvement of their proprietary models. Furthermore, these platforms do not provide the necessary space for public ``deliberation and discourse'', or better even ``dissent, disobedience, and difference'', which the authors consider important constituents of a truly democratic process. What is needed are community-led initiatives that ensure that labour and gains rest with the same group of individuals, and that allow for intervention. 
This directs us to claim (2):  Of course, science will always need some methodological consensus and shared standards to function. In striving for rigour and objectivity, such consensus is best identified among and tested against diverse, conflicting, situated, partial, and changing perspectives. Evaluation should be recognised as a bounded practice, never all-encompassing and general-purpose. Only then can we start to create meaningful measures to evaluate evaluations and assure aptness to context.

 Finally, for epistemic and ethical reasons, the community has to start holding those in power accountable for capturing, exploiting, and gaming evaluation in their own favour. This is a necessary means for mitigating the extent to which the structural injustices in AI benchmarking are avoidable. This paper is a first step in this direction. Instead of vaguely calling for system monitoring through benchmarks, regulators should require true democratic oversight of AI tools when it comes to assessing their suitability and risks. This can only happen through localised, context- and use-specific, as well as continuously recurring modes of evaluation. To this end, regulators must provide the venues and infrastructures needed for its realisation, where diverse stakeholders can meaningfully deliberate and contest decisions.

\section*{Acknowledgements}
This work was supported by the Weizenbaum Institute (grant number 16DII141), funded by the German Federal Ministry of Research, Technology, and Space (BMFTR) and the State of Berlin.

\bibliography{references}

@inproceedings{kraft2025socialbiaspopularquestionanswering,
    title = "Social Bias in Popular Question-Answering Benchmarks",
    author = "Kraft, Angelie  and
      Simon, Judith  and
      Schimmler, Sonja",
    booktitle = "Proceedings of the 14th International Joint Conference on Natural Language Processing and the 4th Conference of the Asia-Pacific Chapter of the Association for Computational Linguistics",
    series = "IJCNLP-AACL '25",
    month = dec,
    year = "2025",
    location = "Mumbai, India",
    publisher = "AFNLP and ACL",
    url = "https://aclanthology.org/2025.ijcnlp-long.79/",
    pages = "1421--1438",
    ISBN = "979-8-89176-298-5",
}

@inproceedings{wallach2024evaluating,
author = {Wallach, Hanna and Desai, Meera and Cooper, A. Feder and Wang, Angelina and Atalla, Chad and Barocas, Solon and Blodgett, Su Lin and Chouldechova, Alexandra and Corvi, Emily and Dow, P. Alex and Garcia-Gathright, Jean and Olteanu, Alexandra and Pangakis, Nicholas and Reed, Stefanie and Sheng, Emily and Vann, Dan and Vaughan, Jennifer Wortman and Vogel, Matthew and Washington, Hannah and Jacobs, Abigail Z.},
title = {Position: Evaluating Generative AI Systems is a Social Science Measurement Challenge},
year = {2025},
publisher = {JMLR.org},
booktitle = {Proceedings of the 42nd International Conference on Machine Learning},
articleno = {3318},
numpages = {20},
location = {Vancouver, Canada},
series = {ICML'25}
}

@inproceedings{powere2024statistical,
  author    = {Powers, Hannah and Baldini, Ioana and Wei, Dennis and  Bennett, Kristin P.},
  title     = {Statistical Bias in Bias Benchmark Design},
    year = {2024},
    publisher = {Curran Associates Inc.},
    address = {Red Hook, NY, USA},
    booktitle = {Proceedings of the 38th International Conference on Neural Information Processing Systems},
    series = {NeurIPS '24},
    location = {Vancouver, BC, Canada}
}

@article{cronbach1955construct,
  title={Construct Validity in Psychological Tests},
  author={Cronbach, Lee J and Meehl, Paul E},
  journal={Psychological Bulletin},
  volume={52},
  number={4},
  pages={281},
  year={1955},
  publisher={American Psychological Association}
}

@inproceedings{
bean2025measuring,
title={Measuring what Matters: Construct Validity in Large Language Model Benchmarks},
author={Andrew M. Bean and Ryan Othniel Kearns and Angelika Romanou and Franziska Sofia Hafner and Harry Mayne and Jan Batzner and Negar Foroutan and Chris Schmitz and Karolina Korgul and Hunar Batra and Oishi Deb and Emma Beharry and Cornelius Emde and Thomas Foster and Anna Gausen and Mar{\'\i}a Grandury and Simeng Han and Valentin Hofmann and Lujain Ibrahim and Hazel Kim and Hannah Rose Kirk and Fangru Lin and Gabrielle Kaili-May Liu and Lennart Luettgau and Jabez Magomere and Jonathan Rystr{\o}m and Anna Sotnikova and Yushi Yang and Yilun Zhao and Adel Bibi and Antoine Bosselut and Ronald Clark and Arman Cohan and Jakob Nicolaus Foerster and Yarin Gal and Scott A. Hale and Inioluwa Deborah Raji and Christopher Summerfield and Philip Torr and Cozmin Ududec and Luc Rocher and Adam Mahdi},
booktitle={The Thirty-ninth Annual Conference on Neural Information Processing Systems Datasets and Benchmarks Track},
series = {NeurIPS '25},
location = {San Diego, CA, USA},
year={2025},
url={https://openreview.net/forum?id=mdA5lVvNcU}
}

@inproceedings{orr2024aisport,
author = {Orr, Will and Kang, Edward B.},
title = {AI as a Sport: On the Competitive Epistemologies of Benchmarking},
year = {2024},
isbn = {9798400704505},
publisher = {ACM},
address = {New York, NY, USA},
url = {https://doi.org/10.1145/3630106.3659012},
doi = {10.1145/3630106.3659012},
booktitle = {Proceedings of the 2024 ACM Conference on Fairness, Accountability, and Transparency},
series = {FAccT '24},
pages = {1875–1884},
numpages = {10},
location = {Rio de Janeiro, Brazil},
}

@inproceedings{raji2021ai,
  author       = {Inioluwa Deborah Raji and
                  Emily Denton and
                  Emily M. Bender and
                  Alex Hanna and
                  Amandalynne Paullada},

  title        = {{AI} and the Everything in the Whole Wide World Benchmark},
  booktitle    = {Proceedings of the Neural Information Processing Systems Track on
                  Datasets and Benchmarks 1, NeurIPS Datasets and Benchmarks 2021},
series = {NeurIPS '21},
location = {Online},
  year         = {2021},
  url          = {https://datasets-benchmarks-proceedings.neurips.cc/paper/2021/hash/084b6fbb10729ed4da8c3d3f5a3ae7c9-Abstract-round2.html},
  bibsource    = {dblp computer science bibliography, https://dblp.org}
}

@inproceedings{parth2024democratic,
author = {sarin, parth and Bao, Michelle},
title = {Democratic Perspectives and Corporate Captures of Crowdsourced Evaluations},
year = {2024},
publisher = {Curran Associates Inc.},
address = {Red Hook, NY, USA},
booktitle = {Proceedings of the 38th International Conference on Neural Information Processing Systems},
series = {NeurIPS '24},
location = {Vancouver, BC, Canada},
}

@inproceedings{bowman2021what,
    title = "What Will it Take to Fix Benchmarking in Natural Language Understanding?",
    author = "Bowman, Samuel R.  and
      Dahl, George",

    booktitle = "Proceedings of the 2021 Conference of the North American Chapter of the Association for Computational Linguistics: Human Language Technologies",
    series = "NAACL '21",
    year = "2021",
    location = "Online",
    publisher = "ACL",
    url = "https://aclanthology.org/2021.naacl-main.385/",
    doi = "10.18653/v1/2021.naacl-main.385",
    pages = "4843--4855"
}

@inproceedings{jacobs-2021-measurement,
author = {Jacobs, Abigail Z. and Wallach, Hanna},
title = {Measurement and Fairness},
year = {2021},
isbn = {9781450383097},
publisher = {ACM},
url = {https://doi.org/10.1145/3442188.3445901},
doi = {10.1145/3442188.3445901},
booktitle = {Proceedings of the 2021 ACM Conference on Fairness, Accountability, and Transparency},
series = {FAccT '21},
pages = {375–385},
numpages = {11},
location = {Online},
address = {New York, NY, USA}, 
}

@misc{uzunoglu2025flaw,
      title={The Flaw of Averages: Quantifying Uniformity of Performance on Benchmarks}, 
      author={Arda Uzunoglu and Tianjian Li and Daniel Khashabi},
      year={2025},
      eprint={2509.25671},
      archivePrefix={arXiv},
      primaryClass={cs.CL},
      url={https://arxiv.org/abs/2509.25671}, 
}

@inproceedings{demchak2024assessing,
  author    = {Demchak, Nathaniel and Guan, Xin and Wu, Zekun and Xu, Ziyi and Koshiyama, Adriano and Kazim, Emre},
  title     = {Assessing Bias in Metric Models for LLM Open-Ended Generation Bias Benchmarks},
  booktitle = {Proceedings of the Workshop Evaluating Evaluations: Examining Best Practices for Measuring Broader Impacts of Generative AI co-located with NeurIPS '24},
location = {Vancouver, BC, Canada},
year = {2024}
}

@article{center2026benchmark,
  title={A Benchmark of Expert-Level Academic Questions to Assess AI Capabilities},
  author={{Center for AI Safety} and {Scale AI} and {HLE Contributors Consortium}},
  journal={Nature},
  volume={649},
  number={8099},
  pages={1139--1146},
  year={2026},
  publisher={Nature Publishing Group},
  address={UK London}
}

@book{bowker1999sorting,
    author = {Bowker, Geoffrey C. and Star, Susan Leigh},
    title = {Sorting Things Out: Classification and Its Consequences},
    publisher = {The MIT Press},
    year = {1999},
    month = {09},
    isbn = {9780262269070},
    doi = {10.7551/mitpress/6352.001.0001},
    url = {https://doi.org/10.7551/mitpress/6352.001.0001},
}

@article{aiact,
    author ={{European Union}},
title = {{Regulation (EU) 2024/1689 – Regulation (EU) 2024/1689 of the European Parliament and of the Council of 13 June 2024 laying down harmonised rules on artificial intelligence and amending Regulations (EC) No 300/2008, (EU) No 167/2013, (EU) No 168/2013, (EU) 2018/858, (EU) 2018/1139 and (EU) 2019/2144 and Directives 2014/90/EU, (EU) 2016/797 and (EU) 2020/1828 (Artificial Intelligence Act) (Text with EEA relevance)}},
journal = {Official Journal
of the European Union},
year = {2024},
}

@article{bommasani2023,
author = {Bommasani, Rishi and Liang, Percy and Lee, Tony},
title = {Holistic Evaluation of Language Models},
journal = {Annals of the New York Academy of Sciences},
volume = {1525},
number = {1},
pages = {140-146},
doi = {https://doi.org/10.1111/nyas.15007},
url = {https://nyaspubs.onlinelibrary.wiley.com/doi/abs/10.1111/nyas.15007},
eprint = {https://nyaspubs.onlinelibrary.wiley.com/doi/pdf/10.1111/nyas.15007},
year = {2023},
address = {New York, NY, USA}
}

@article{frontiermath,
  author       = {Elliot Glazer and
                  Ege Erdil and
                  Tamay Besiroglu and
                  Diego Chicharro and
                  Evan Chen and
                  Alex Gunning and
                  Caroline Falkman Olsson and
                  Jean{-}Stanislas Denain and
                  Anson Ho and
                  Emily de Oliveira Santos and
                  Olli J{\"{a}}rviniemi and
                  Matthew Barnett and
                  Robert Sandler and
                  Matej Vrzala and
                  Jaime Sevilla and
                  Qiuyu Ren and
                  Elizabeth Pratt and
                  Lionel Levine and
                  Grant Barkley and
                  Natalie Stewart and
                  Bogdan Grechuk and
                  Tetiana Grechuk and
                  Shreepranav Varma Enugandla and
                  Mark Wildon},
  title        = {FrontierMath: {A} Benchmark for Evaluating Advanced Mathematical Reasoning
                  in {AI}},
  journal      = {CoRR},
  volume       = {abs/2411.04872},
  year         = {2024},
  url          = {https://doi.org/10.48550/arXiv.2411.04872},
  doi          = {10.48550/ARXIV.2411.04872},
  eprinttype   = {arXiv},
  eprint       = {2411.04872},
  bibsource    = {dblp computer science bibliography, https://dblp.org}
}

@inproceedings{realtoxicityprompts,
  author       = {Samuel Gehman and
                  Suchin Gururangan and
                  Maarten Sap and
                  Yejin Choi and
                  Noah A. Smith},

  title        = {RealToxicityPrompts: Evaluating Neural Toxic Degeneration in Language
                  Models},
  booktitle    = {Findings of the Association for Computational Linguistics: {EMNLP}
                  2020},
location = {Online},
  pages        = {3356--3369},
  publisher    = {ACL},
  year         = {2020},
  url          = {https://doi.org/10.18653/v1/2020.findings-emnlp.301},
  doi          = {10.18653/V1/2020.FINDINGS-EMNLP.301},
  bibsource    = {dblp computer science bibliography, https://dblp.org}
}

@inproceedings{liu-etal-2024-ecbd,
    title = "{ECBD}: Evidence-Centered Benchmark Design for {NLP}",
    author = "Liu, Yu Lu  and
      Blodgett, Su Lin  and
      Cheung, Jackie  and
      Liao, Q. Vera  and
      Olteanu, Alexandra  and
      Xiao, Ziang",
    booktitle = "Proceedings of the 62nd Annual Meeting of the Association for Computational Linguistics (Volume 1: Long Papers)",
    series = "ACL '24",
    year = "2024",
    location = "Bangkok, Thailand",
    publisher = "ACL",
    url = "https://aclanthology.org/2024.acl-long.861/",
    doi = "10.18653/v1/2024.acl-long.861",
    pages = "16349--16365"
}

@book{young-1990-justice,
    author = {Young, Iris Marion},
    title = {Justice and the Politics of Difference},
    publisher = {Princeton University Press},
    year = {1990}
}

@book{young-2010-responsibility,
    author = {Young, Iris Marion},
    title = {Responsibility for Justice},
    publisher = {Oxford University Press},
    year = {2010}
}

@misc{bordes2025evalfactsheetsstructuredframework,
      title={Eval Factsheets: A Structured Framework for Documenting AI Evaluations}, 
      author={Florian Bordes and Candace Ross and Justine T Kao and Evangelia Spiliopoulou and Adina Williams},
      year={2025},
      eprint={2512.04062},
      archivePrefix={arXiv},
      primaryClass={cs.LG},
      url={https://arxiv.org/abs/2512.04062}, 
}

@article{gebru2021datasheets,
author = {Gebru, Timnit and Morgenstern, Jamie and Vecchione, Briana and Vaughan, Jennifer Wortman and Wallach, Hanna and III, Hal Daum\'{e} and Crawford, Kate},
title = {Datasheets for datasets},
year = {2021},
issue_date = {December 2021},
publisher = {ACM},
address = {New York, NY, USA},
volume = {64},
number = {12},
issn = {0001-0782},
url = {https://doi.org/10.1145/3458723},
doi = {10.1145/3458723},
journal = {Commun. ACM},
month = nov,
pages = {86–92},
numpages = {7}
}

@article{Young_2006,
author={Young, Iris Marion},
title={Responsibility and Global Justice: A Social Connection Model},
year={2006},
journal={Social Philosophy and Policy},
volume={23}, 
number={1},
DOI={10.1017/S0265052506060043}, 
pages={102–130}
}

@article{McKeown_2021,
author = {McKeown, Maeve},
title = {Structural Injustice},
journal = {Philosophy Compass},
volume = {16},
number = {7},
pages = {e12757},
doi = {https://doi.org/10.1111/phc3.12757},
url = {https://compass.onlinelibrary.wiley.com/doi/abs/10.1111/phc3.12757},
eprint = {https://compass.onlinelibrary.wiley.com/doi/pdf/10.1111/phc3.12757},
year = {2021}
}

@book{McKeown_2024,
author = {McKeown, Maeve},
title = {With Power Comes Responsibility: The Politics of Structural Injustice},
Publisher = {Bloomsbury},
year = {2024}
}

@inproceedings{Jain_2024_algorithmicpluralism, 
author = {Jain, Shomik and Suriyakumar, Vinith and Creel, Kathleen and Wilson, Ashia}, 
title = {Algorithmic Pluralism: A Structural Approach To Equal Opportunity}, 
year = {2024}, 
booktitle = {Proceedings of the 2024 ACM Conference on Fairness, Accountability, and Transparency},
isbn = {9798400704505}, publisher = {ACM}, 
address = {New York, NY, USA}, 
url = {https://doi.org/10.1145/3630106.3658899}, 
doi = {10.1145/3630106.3658899}, 
pages = {197–206}, 
numpages = {10}, 
location = {Rio de Janeiro, Brazil}, 
series = {FAccT '24} 
}

@inproceedings{Bommasani_2022_algorithmicmonoculture,
 author = {Bommasani, Rishi and Creel, Kathleen A. and Kumar, Ananya and Jurafsky, Dan and Liang, Percy S},
 booktitle = {Advances in Neural Information Processing Systems},
 editor = {S. Koyejo and S. Mohamed and A. Agarwal and D. Belgrave and K. Cho and A. Oh},
 pages = {3663--3678},
 publisher = {Curran Associates Inc.},
address = {Red Hook, NY, USA},
 title = {Picking on the Same Person: Does Algorithmic Monoculture lead to Outcome Homogenization?},
 url = {https://proceedings.neurips.cc/paper_files/paper/2022/file/17a234c91f746d9625a75cf8a8731ee2-Paper-Conference.pdf},
 volume = {35},
 year = {2022}
}

@inproceedings{Naudts_2024,
author = {Naudts, Laurens}, 
title = {The Digital Faces of Oppression and Domination: A Relational and Egalitarian Perspective on the Data-driven Society and its Regulation}, 
year = {2024}, 
isbn = {9798400704505}, 
publisher = {ACM}, 
address = {New York, NY, USA}, 
url = {https://doi.org/10.1145/3630106.3658934}, 
doi = {10.1145/3630106.3658934}, 
booktitle = {Proceedings of the 2024 ACM Conference on Fairness, Accountability, and Transparency}, 
series = {FAccT '24},
pages = {701–712}, 
numpages = {12}, 
location = {Rio de Janeiro, Brazil}, 
}

@inproceedings{koch2021reduced,
  author = {Koch, Bernard and Denton, Emily and Hanna, Alex and Foster, Jacob G},
  booktitle    = {Proceedings of the Neural Information Processing Systems Track on
                  Datasets and Benchmarks 1, NeurIPS Datasets and Benchmarks 2021},
location = {Online},
 title = {Reduced, Reused and Recycled: The Life of a  Dataset in Machine Learning Research},
 url = {https://datasets-benchmarks-proceedings.neurips.cc/paper/2021/file/3b8a614226a953a8cd9526fca6fe9ba5-Paper-round2.pdf},
series = {NeurIPS '21},
 year = {2021}
}

@inproceedings{birhane2022power,
author = {Birhane, Abeba and Isaac, William and Prabhakaran, Vinodkumar and Diaz, Mark and Elish, Madeleine Clare and Gabriel, Iason and Mohamed, Shakir},
title = {Power to the People? Opportunities and Challenges for Participatory AI},
year = {2022},
isbn = {9781450394772},
publisher = {ACM},
address = {New York, NY, USA},
url = {https://doi.org/10.1145/3551624.3555290},
doi = {10.1145/3551624.3555290},
booktitle = {Proceedings of the 2nd ACM Conference on Equity and Access in Algorithms, Mechanisms, and Optimization},
articleno = {6},
numpages = {8},
location = {Arlington, VA, USA},
series = {EAAMO '22}
}

@article{couldry2019datacolonialism,
author = {Nick Couldry and Ulises A. Mejias},
title ={Data Colonialism: Rethinking Big Data’s Relation to the Contemporary Subject},
journal = {Television \& New Media},
volume = {20},
number = {4},
pages = {336-349},
year = {2019},
doi = {10.1177/1527476418796632},
URL = { 
        https://doi.org/10.1177/1527476418796632
},
eprint = { 
    
        https://doi.org/10.1177/1527476418796632
}
}

@inproceedings{kapania2026survival,
author = {Kapania, Shivani and Yang, Tianling and Abdelkadir, Nuredin Ali and Scheuerman, Morgan Klaus and Miceli, Milagros and Taylor, Alex S and Fox, Sarah E},
title = { 'The plan is just survival': Data Work in Kenya and the Regime of Entrapment},
year = {2026},
isbn = {9798400722783},
publisher = {ACM},
address = {New York, NY, USA},
url = {https://doi.org/10.1145/3772318.3791097},
doi = {10.1145/3772318.3791097},
booktitle = {Proceedings of the 2026 CHI Conference on Human Factors in Computing Systems},
articleno = {936},
numpages = {17},
location = {
},
series = {CHI '26}
}

@inproceedings{gebrekidan2024contentmoderation,
    author = {Gebrekidan, F. B.},
year = {2024},
title = {Content Moderation: The Harrowing, Traumatizing Job that Left Many African Data Workers with Mental Health Issues and Drug Dependency},
booktitle = {Data Workers‘ Inquiry},
editors = {M. Miceli and A. Dinika and K. Kauffman and C. Salim Wagner and L. Sachenbacher},
url = {https://data-workers.org/fasica}
}

@inproceedings{
schaeffer2025causally,
title={Causally Quantifying the Effect of Test Set Contamination on Generative Benchmarks},
author={Rylan Schaeffer and Brando Miranda and Joshua Kazdan and Ken Liu and Ahmed M Ahmed and Niloofar Mireshghallah and Sanmi Koyejo},
booktitle={NeurIPS '25 Workshop on Evaluating the Evolving LLM Lifecycle: Benchmarks, Emergent Abilities, and Scaling},
location = {San Diego, CA, USA},
series = {NeurIPS '25},
year={2025},
url={https://openreview.net/forum?id=RsmjshBEDP}
}

@inproceedings{
sainz2023nlp,
title={{NLP} Evaluation in trouble: On the Need to Measure {LLM} Data Contamination for each Benchmark},
author={Oscar Sainz and Jon Ander Campos and Iker Garc{\'\i}a-Ferrero and Julen Etxaniz and Oier Lopez de Lacalle and Eneko Agirre},
booktitle={The 2023 Conference on Empirical Methods in Natural Language Processing},
series = {EMNLP '23},
year={2023},
url={https://openreview.net/forum?id=KivNpBsfAS}
}

@inproceedings{jimenez2024swebench,
 author = {Jimenez, Carlos E and Yang, John and Wettig, Alexander and Yao, Shunyu and Pei, Kexin and Press, Ofir and Narasimhan, Karthik},
 booktitle = {International Conference on Learning Representations},
 editor = {B. Kim and Y. Yue and S. Chaudhuri and K. Fragkiadaki and M. Khan and Y. Sun},
 pages = {54107--54157},
 title = {SWE-bench: Can Language Models Resolve Real-world Github Issues?},
 url = {https://proceedings.iclr.cc/paper_files/paper/2024/file/edac78c3e300629acfe6cbe9ca88fb84-Paper-Conference.pdf},
 series = {ICLR '24},
 year = {2024}
}

@misc{hartmann2026byebyeperspectiveapi,
      title={Bye Bye Perspective API: Lessons for Measurement Infrastructure in NLP, CSS and LLM Evaluation}, 
      author={David Hartmann and Manuel Tonneau and Angelie Kraft and LK Seiling and Dimitri Staufer and Pieter Delobelle and Jan Fillies and Anna Ricarda Luther and Jan Batzner and Mareike Lisker},
      year={2026},
      eprint={2604.25580},
      archivePrefix={arXiv},
      primaryClass={cs.CL},
      url={https://arxiv.org/abs/2604.25580}, 
}

@inproceedings{marin2026aimodelaccurateenough,
author = {Uberti-Bona Marin, Lucas Giovanni and Rijsbosch, Bram and Meding, Kristof and Spanakis, Gerasimos and van Dijck, Gijs and Kollnig, Konrad},
title = {Is your AI Model Accurate Enough? The Difficult Choices Behind Rigorous AI Development and the EU AI Act},
year = {2026},
isbn = {9798400725968},
publisher = {ACM},
address = {New York, NY, USA},
url = {https://doi.org/10.1145/3805689.3806436},
doi = {10.1145/3805689.3806436},
booktitle = {Proceedings of the 2026 ACM Conference on Fairness, Accountability, and Transparency},
pages = {8184–8201},
numpages = {18},
location = {
},
series = {FAccT '26}
}

@inproceedings{umutlu-etal-2025-evaluating,
    title = "Evaluating the Quality of Benchmark Datasets for Low-Resource Languages: A Case Study on {T}urkish",
    author = "Umutlu, Elif Ecem  and
      Cengiz, Ayse Aysu  and
      Sever, Ahmet Kaan  and
      Erdem, Seyma  and
      Aytan, Burak  and
      Tufan, Busra  and
      Topraksoy, Abdullah  and
      Dar{\i}c{\i}, Esra  and
      Toraman, Cagri",
    booktitle = "Proceedings of the Fourth Workshop on Generation, Evaluation and Metrics", 
    series = "GEM{\texttwosuperior}",
    month = jul,
    year = "2025",
    location = "Vienna, Austria and virtual meeting",
    publisher = "ACL",
    url = "https://aclanthology.org/2025.gem-1.41/",
    pages = "471--487",
    ISBN = "979-8-89176-261-9"
}

@article{miceli2022dataproduction,
author = {Miceli, Milagros and Posada, Julian},
title = {The Data-Production Dispositif},
year = {2022},
issue_date = {November 2022},
publisher = {ACM},
address = {New York, NY, USA},
volume = {6},
number = {CSCW2},
url = {https://doi.org/10.1145/3555561},
doi = {10.1145/3555561},
journal = {Proc. ACM Hum.-Comput. Interact.},
month = nov,
articleno = {460},
numpages = {37}
}

@inproceedings{blodgett2020language,
    title = "Language (Technology) is Power: A Critical Survey of {``}Bias{''} in {NLP}",
    author = "Blodgett, Su Lin  and
      Barocas, Solon  and
      Daum{\'e} III, Hal  and
      Wallach, Hanna",
    booktitle = "Proceedings of the 58th Annual Meeting of the Association for Computational Linguistics",
    series = "ACL '20",
    year = "2020",
    location = "Online",
    publisher = "ACL",
    url = "https://aclanthology.org/2020.acl-main.485",
    doi = "10.18653/v1/2020.acl-main.485",
    pages = "5454--5476",
}

@article{dotson2014conceptualizing,
author = {Kristie Dotson},
title = {Conceptualizing Epistemic Oppression},
journal = {Social Epistemology},
volume = {28},
number = {2},
pages = {115--138},
year = {2014},
publisher = {Routledge},
doi = {10.1080/02691728.2013.782585},
URL = { https://doi.org/10.1080/02691728.2013.782585
},
eprint = {https://doi.org/10.1080/02691728.2013.782585
}
}

@article{dotson2012cautionarytale,
 ISSN = {01609009, 15360334},
 URL = {http://www.jstor.org/stable/10.5250/fronjwomestud.33.1.0024},
 author = {Kristie Dotson},
 journal = {Frontiers: A Journal of Women Studies},
 number = {1},
 pages = {24--47},
 publisher = {[University of Nebraska Press, Frontiers, Inc.]},
 title = {A Cautionary Tale: On Limiting Epistemic Oppression},
 urldate = {2025-10-31},
 volume = {33},
 year = {2012}
}

@phdthesis{kraft2025knowledge,
  title={On Knowledge in AI: Epistemic and Ethical Limitations of Language Models and Knowledge Graphs},
  author={Kraft, Angelie},
  year={2025},
  school={University of Hamburg, Hamburg},
  url={https://ediss.sub.uni-hamburg.de/handle/ediss/12231}
}

@book{truong2026measurement,
    author = {Truong, S. T. and Koyejo, S.},
    title = {AI Measurement Science: A Science of Knowing Where AI Thrives, Where It Breaks, and How to Respond},
    publisher = {Stanford University},
    year = {2026}
}

@inproceedings{chiang2024chatbotarena,
author = {Chiang, Wei-Lin and Zheng, Lianmin and Sheng, Ying and Angelopoulos, Anastasios N. and Li, Tianle and Li, Dacheng and Zhu, Banghua and Zhang, Hao and Jordan, Michael I. and Gonzalez, Joseph E. and Stoica, Ion},
title = {Chatbot Arena: An Open Platform for Evaluating LLMs by Human Preference},
year = {2024},
publisher = {JMLR.org},
booktitle = {Proceedings of the 41st International Conference on Machine Learning},
articleno = {331},
numpages = {30},
location = {Vienna, Austria},
series = {ICML '24}
}

@misc{ganguli2022redteaming,
title={Red Teaming Language Models to Reduce Harms: Methods, Scaling Behaviors, and Lessons Learned}, 
      author={Deep Ganguli and Liane Lovitt and Jackson Kernion and Amanda Askell and Yuntao Bai and Saurav Kadavath and Ben Mann and Ethan Perez and Nicholas Schiefer and Kamal Ndousse and Andy Jones and Sam Bowman and Anna Chen and Tom Conerly and Nova DasSarma and Dawn Drain and Nelson Elhage and Sheer El-Showk and Stanislav Fort and Zac Hatfield-Dodds and Tom Henighan and Danny Hernandez and Tristan Hume and Josh Jacobson and Scott Johnston and Shauna Kravec and Catherine Olsson and Sam Ringer and Eli Tran-Johnson and Dario Amodei and Tom Brown and Nicholas Joseph and Sam McCandlish and Chris Olah and Jared Kaplan and Jack Clark},
      year={2022},
      eprint={2209.07858},
      archivePrefix={arXiv},
      primaryClass={cs.CL},
      url={https://arxiv.org/abs/2209.07858}, 
}

@book{Longino2002,
url = {https://doi.org/10.1515/9780691187013},
title = {The Fate of Knowledge},
author = {Longino, Helen},
publisher = {Princeton University Press},
doi = {doi:10.1515/9780691187013},
isbn = {9780691187013},
year = {2002},
lastchecked = {2023-11-17}
}

@incollection{haraway2016situated,
  title={Situated Knowledges: The Science Question in Feminism and the Privilege of Partial Perspective},
  author={Haraway, Donna},
  booktitle={Space, Gender, Knowledge: Feminist Readings},
  pages={53--72},
  year={2016},
  publisher={Routledge},
  url = {https://www.jstor.org/stable/3178066}
}

@inproceedings{longpre2025bridging,
  author       = {Shayne Longpre and
                  Nikhil Singh and
                  Manuel Cherep and
                  Kushagra Tiwary and
                  Joanna Materzynska and
                  William Brannon and
                  Robert Mahari and
                  Naana Obeng{-}Marnu and
                  Manan Dey and
                  Mohammed Hamdy and
                  Nayan Saxena and
                  Ahmad Mustafa Anis and
                  Emad A. Alghamdi and
                  Vu Minh Chien and
                  Da Yin and
                  Kun Qian and
                  Yizhi Li and
                  Minnie Liang and
                  An Dinh and
                  Shrestha Mohanty and
                  et al.},
  title        = {Bridging the Data Provenance Gap Across Text, Speech, and Video},
  booktitle    = {The Thirteenth International Conference on Learning Representations},
  series = {ICLR '25},
  publisher    = {OpenReview.net},
  year         = {2025},
  url          = {https://openreview.net/forum?id=G5DziesYxL},
  bibsource    = {dblp computer science bibliography, https://dblp.org}
}

@incollection{Herzog2021AlgorithmicBias,
  author    = {Herzog, Lisa},
  title     = {Algorithmic Bias and Access to Opportunities},
  editor    = {V{\'e}liz, Carissa},
  booktitle = {The Oxford Handbook of Digital Ethics},
  publisher = {Oxford University Press},
  pages     = {413--432},
  year      = {2021},
  doi       = {10.1093/oxfordhb/9780198857815.013.21},
  isbn      = {9780198857815}
}

@incollection{Browne2023AIStructuralInjustice,
  author    = {Browne, Jude},
  title     = {AI and Structural Injustice: A Feminist Perspective},
  editor    = {Browne, Jude and Cave, Stephen and Drage, Eleanor and McInerney, Kerry},
  booktitle = {Feminist AI: Critical Perspectives on Algorithms, Data, and Intelligent Machines},
  publisher = {Oxford University Press},
  pages     = {328--346},
  year      = {2023},
  doi       = {10.1093/oso/9780192889898.003.0019},
  isbn      = {9780192889898}
}

@inproceedings{Kasirzadeh2022AlgorithmicFairness,
  author    = {Kasirzadeh, Atoosa},
  title     = {Algorithmic Fairness and Structural Injustice: Insights from Feminist Political Philosophy},
  booktitle = {Proceedings of the 2022 AAAI/ACM Conference on AI, Ethics, and Society},
  series    = {AIES '22},
  pages     = {349--356},
  publisher = {ACM},
  address   = {New York, NY, USA},
  year      = {2022},
  doi       = {10.1145/3514094.3534188},
  isbn      = {9781450392471}
}

@article{HerzogBranford2025RelationalEthics,
  author  = {Herzog, Christian and Branford, Jason},
  title   = {Relational Ethics and Structural Epistemic Injustice of AI in Medicine},
  journal = {Philosophy \& Technology},
  volume  = {38},
  number  = {4},
  pages   = {160},
  year    = {2025},
  doi     = {10.1007/s13347-025-00987-1}
}

@book{Hao2025EmpireOfAI,
  author    = {Hao, Karen},
  title     = {Empire of AI: Dreams and Nightmares in Sam Altman's OpenAI},
  publisher = {Penguin Press},
  address   = {New York},
  year      = {2025},
  isbn      = {9780593657508}
}

@book{Crawford2021AtlasOfAI,
  author    = {Crawford, Kate},
  title     = {Atlas of AI: Power, Politics, and the Planetary Costs of Artificial Intelligence},
  publisher = {Yale University Press},
  address   = {New Haven, CT},
  year      = {2021},
  isbn      = {9780300209570}
}

@article{Ott2022MappingGlobalDynamics,
  author  = {Ott, Simon and Barbosa-Silva, Adriano and Blagec, Kathrin and Brauner, Jan and Samwald, Matthias},
  title   = {Mapping Global Dynamics of Benchmark Creation and Saturation in Artificial Intelligence},
  journal = {Nature Communications},
  volume  = {13},
  pages   = {6793},
  year    = {2022},
  doi     = {10.1038/s41467-022-34591-0}
}

@article{ThomasUminsky2022RelianceOnMetrics,
  author  = {Thomas, Rachel L. and Uminsky, David},
  title   = {Reliance on Metrics Is a Fundamental Challenge for AI},
  journal = {Patterns},
  volume  = {3},
  number  = {5},
  pages   = {100476},
  year    = {2022},
  doi     = {10.1016/j.patter.2022.100476}
}

@book{MuldoonGrahamCant2024FeedingTheMachine,
  author    = {Muldoon, James and Graham, Mark and Cant, Callum},
  title     = {Feeding the Machine: The Hidden Human Labour Powering AI},
  publisher = {Canongate Books},
  address   = {Edinburgh},
  year      = {2024},
  isbn      = {9781838859114}
}

@inproceedings{Strubell2019EnergyPolicy,
  author    = {Strubell, Emma and Ganesh, Ananya and McCallum, Andrew},
  title     = {Energy and Policy Considerations for Deep Learning in {NLP}},
  booktitle = {Proceedings of the 57th Annual Meeting of the Association for Computational Linguistics},
  series = {ACL '19},
  pages     = {3645--3650},
  publisher = {ACL},
  location   = {Florence, Italy},
  year      = {2019},
  doi       = {10.18653/v1/P19-1355}
}

@inproceedings{Bender2021StochasticParrots,
  author    = {Bender, Emily M. and Gebru, Timnit and McMillan-Major, Angelina and Shmitchell, Shmargaret},
  title     = {On the Dangers of Stochastic Parrots: Can Language Models Be Too Big?},
  booktitle = {Proceedings of the 2021 ACM Conference on Fairness, Accountability, and Transparency},
  series = {FAccT '21},
  pages     = {610--623},
  publisher = {ACM},
  address   = {New York, NY, USA},
  year      = {2021},
  doi       = {10.1145/3442188.3445922},
  isbn      = {9781450383097}
}

@article{Ricaurte2019DataEpistemologies,
  author  = {Ricaurte, Paola},
  title   = {Data Epistemologies, the Coloniality of Power, and Resistance},
  journal = {Television \& New Media},
  volume  = {20},
  number  = {4},
  pages   = {350--365},
  year    = {2019},
  doi     = {10.1177/1527476419831640}
}

@article{Ricaurte2022EthicsMajorityWorld,
  author  = {Ricaurte, Paola},
  title   = {Ethics for the Majority World: {AI} and the Question of Violence at Scale},
  journal = {Media, Culture \& Society},
  volume  = {44},
  number  = {4},
  pages   = {726--745},
  year    = {2022},
  doi     = {10.1177/01634437221099612}
}

@article{MohamedPngIsaac2020DecolonialAI,
  author  = {Mohamed, Shakir and Png, Marie-Therese and Isaac, William},
  title   = {Decolonial {AI}: Decolonial Theory as Sociotechnical Foresight in Artificial Intelligence},
  journal = {Philosophy \& Technology},
  volume  = {33},
  pages   = {659--684},
  year    = {2020},
  doi     = {10.1007/s13347-020-00405-8}
}

@article{Cartwright2006WellOrderedScience,
  author  = {Cartwright, Nancy},
  title   = {Well-Ordered Science: Evidence for Use},
  journal = {Philosophy of Science},
  volume  = {73},
  number  = {5},
  pages   = {981--990},
  year    = {2006},
  doi     = {10.1086/518803}
}

@book{Kitcher2001ScienceTruthDemocracy,
  author    = {Kitcher, Philip},
  title     = {Science, Truth, and Democracy},
  publisher = {Oxford University Press},
  address   = {New York},
  year      = {2001},
  doi       = {10.1093/0195145836.001.0001},
  isbn      = {9780195145830}
}

@book{Kitcher2011ScienceDemocraticSociety,
  author    = {Kitcher, Philip},
  title     = {Science in a Democratic Society},
  publisher = {Prometheus Books},
  address   = {Amherst, NY},
  year      = {2011},
  isbn      = {9781616144074}
}

@inproceedings{Sloane2022ParticipationNotDesignFix,
  author    = {Sloane, Mona and Moss, Emanuel and Awomolo, Olaitan and Forlano, Laura},
  title     = {Participation Is not a Design Fix for Machine Learning},
  booktitle = {Proceedings of the 2nd {ACM} Conference on Equity and Access in Algorithms, Mechanisms, and Optimization},
  series    = {EAAMO '22},
  pages     = {1--6},
  publisher = {ACM},
  address   = {New York, NY, USA},
  year      = {2022},
  doi       = {10.1145/3551624.3555285}
}

@inproceedings{baack2026unsteadymetricsbenchmarkingcultures,
author = {Baack, Stefan and Buschek, Christo and Bohacek, Maty},
title = {Unsteady Metrics and Benchmarking Cultures of AI Model Builders},
year = {2026},
isbn = {9798400725968},
publisher = {ACM},
address = {New York, NY, USA},
url = {https://doi.org/10.1145/3805689.3812240},
doi = {10.1145/3805689.3812240},
booktitle = {Proceedings of the 2026 ACM Conference on Fairness, Accountability, and Transparency},
pages = {840–872},
numpages = {33},
location = {
},
series = {FAccT '26}
}

@article{Branford2025GenerativeAI,
  author  = {Branford, Jason and Soulier, Elo{\"i}se and Fichtner, Laura},
  title   = {Generative {AI} and Democratic Culture},
  journal = {Philosophy \& Technology},
  volume  = {38},
  number  = {123},
  year    = {2025},
  doi     = {10.1007/s13347-025-00953-x}
}


\end{document}